\documentclass[pdflatex,sn-mathphys-num]{sn-jnl}

\usepackage{graphicx}
\usepackage{multirow}
\usepackage{amsmath,amssymb,amsfonts}
\usepackage{amsthm}
\usepackage{mathrsfs}
\usepackage[title]{appendix}
\usepackage{xcolor}
\usepackage{textcomp}
\usepackage{manyfoot}
\usepackage{booktabs}
\usepackage{algorithm}
\usepackage{algorithmicx}
\usepackage{algpseudocode}
\usepackage{listings}

\theoremstyle{thmstyleone}

\theoremstyle{thmstyletwo}

\theoremstyle{thmstylethree}

\begin{document}

\title{Evaluating Supervised Machine Learning Models: Principles, Pitfalls, and Metric Selection}

\author[1,4]{\fnm{Xuanyan} \sur{Liu}}
\author[2]{\fnm{Ignacio} \sur{Cabrera Martin}}
\author*[3]{\fnm{Marcello} \sur{Trovati}}
\author[1]{\fnm{Xiaolong} \sur{Xu}}
\author[2]{\fnm{Nikolaos} \sur{Polatidis}}

\affil[1]{\orgname{Jiangsu Key Laboratory of Big Data Security and Intelligent Processing, Nanjing University of Posts and Telecommunications}, \orgaddress{\city{Nanjing}, \postcode{210023}, \country{China}}}
\affil[2]{\orgname{University of Brighton}, \country{United Kingdom}}
\affil[3]{\orgname{University of Lancashire}, \country{United Kingdom}}
\affil[4]{\orgname{School of Cyber Science and Engineering, Southeast University}, \orgaddress{\city{Nanjing}, \postcode{211189}, \country{China}}}

\abstract{The evaluation of supervised machine learning models is a critical stage in the development of reliable predictive systems. Despite the widespread availability of machine learning libraries and automated workflows, model assessment is often reduced to the reporting of a small set of aggregate metrics, which can lead to misleading conclusions about real-world performance. This paper examines the principles, challenges, and practical considerations involved in evaluating supervised learning algorithms across classification and regression tasks. In particular, it discusses how evaluation outcomes are influenced by dataset characteristics, validation design, class imbalance, asymmetric error costs, and the choice of performance metrics. Through a series of controlled experimental scenarios using diverse benchmark datasets, the study highlights common pitfalls such as the accuracy paradox, data leakage, inappropriate metric selection, and overreliance on scalar summary measures. The paper also compares alternative validation strategies and emphasizes the importance of aligning model evaluation with the intended operational objective of the task. By presenting evaluation as a decision-oriented and context-dependent process, this work provides a structured foundation for selecting metrics and validation protocols that support statistically sound, robust, and trustworthy supervised machine learning systems.}


\maketitle

\section{Introduction}

The evaluation of supervised machine learning (ML) models is a central component of the model development lifecycle, yet it is often treated as a routine or purely procedural step. Although modern software libraries have made model training increasingly accessible, the reliable assessment of predictive performance remains challenging. In practice, models that achieve excellent results during offline testing may still fail after deployment because the evaluation protocol does not reflect real-world conditions or because issues such as data leakage and distribution shift are overlooked \cite{sculley_hidden_2015, kapoor_leakage_2023}. This gap between apparent validation success and operational performance highlights the need for more rigorous and context-aware evaluation methods.

The primary objective of model evaluation is to estimate how well a learned model generalizes to unseen data \cite{hastie_elements_2009}. However, generalization cannot be reduced to a single universal criterion. Its measurement depends on the relationship between the evaluation strategy, the structure and quality of the dataset, and the practical requirements of the application domain. Across classification, regression, and ranking tasks, model performance is inherently multidimensional. As a result, relying on a single metric can produce misleading conclusions. For example, accuracy may be highly deceptive in imbalanced classification settings, while $R^2$ alone may conceal systematic regression errors that become visible only through residual analysis \cite{chicco_advantages_2020, hand_measuring_2009}.

This paper presents a critical examination of the core principles involved in evaluating supervised learning algorithms through a set of controlled experimental scenarios. Rather than limiting the discussion to standard evaluation procedures, we investigate several common sources of misinterpretation and error, including the choice between hold-out validation and cross-validation under non-IID (Independent and Identically Distributed) conditions \cite{arlot_survey_2010}, the effects of data leakage \cite{kapoor_leakage_2023}, and the trade-offs associated with different performance metrics. Particular attention is given to the accuracy trap in imbalanced datasets, the impact of asymmetric misclassification costs, and the sensitivity of regression measures to outliers and shifts in baseline distributions. By treating evaluation as a decision-oriented and domain-sensitive process, this work aims to support the development of machine learning systems that are not only statistically reliable, but also robust, interpretable, and trustworthy in practice.

The main contributions of this paper are as follows:

\begin{itemize}
    \item We provide a structured overview of the key principles underlying the evaluation of supervised machine learning models.
    \item We analyze the strengths and limitations of common evaluation strategies, including hold-out validation and cross-validation, particularly in non-IID settings.
    \item We examine frequent evaluation pitfalls such as data leakage, misleading metrics under class imbalance, and improper interpretation of regression performance.
    \item We compare the behavior of different performance metrics under controlled experimental scenarios, highlighting their sensitivity to dataset characteristics and task-specific objectives.
    \item We frame model evaluation as a decision-theoretic and application-dependent process, emphasizing its importance for building reliable and operationally trustworthy ML systems.
\end{itemize}

The remainder of this paper is organized as follows. Section \ref{relatedwork} reviews the related work.  Section \ref{problemstatement} defines the problem. Section \ref{evaluationmetrics} discusses the evaluation metrics considered in this study. Section \ref{datasets} describes the datasets used in the experiments. Section \ref{evaluation} presents the experimental evaluation, and Section \ref{conclusions} concludes the paper.

\section{Related work}
\label{relatedwork}

The evaluation of supervised machine learning models has been studied across several interconnected research areas, including statistical learning, model selection, forecasting, uncertainty quantification, and domain-specific validation methodologies. A foundational reference in this space is \textit{The Elements of Statistical Learning}, which identifies generalization, overfitting, resampling, and the bias--variance trade-off as core issues in predictive modeling \cite{hastie_elements_2009}. Within this broader context, cross-validation has emerged as one of the principal tools for model assessment and selection. Arlot and Celisse provide a comprehensive survey of cross-validation methods, clarifying the conditions under which different variants are appropriate for estimating predictive risk and supporting model choice \cite{arlot_survey_2010}. Subsequent work has extended these discussions through theoretical analysis of hold-out design \cite{mcalinn_optimal_2025} and broader reflections on how to establish credible predictive evidence in applied research settings \cite{poldrack_establishment_2020}.

A major theme in the literature is that unreliable evaluation often stems not only from poor model choice, but from weaknesses in experimental design. Sculley et al.\ argue that machine learning systems accumulate hidden technical debt when development and validation shortcuts obscure how models will behave in deployment \cite{sculley_hidden_2015}. Kapoor and Narayanan further demonstrate that data leakage can systematically inflate empirical results and contribute to irreproducible conclusions in machine-learning-based scientific research \cite{kapoor_leakage_2023}. These studies collectively show that evaluation cannot be separated from issues such as data provenance, preprocessing rigor, and the integrity of the overall modeling pipeline.

Another important line of research concerns the choice of evaluation metric. In classification tasks, accuracy remains widely used, but it has long been criticized for its inability to reflect class imbalance and asymmetric error costs. Hand questions the general validity of the AUC as a universal evaluation criterion, showing that its interpretation depends on implicit and potentially inconsistent assumptions about misclassification costs \cite{hand_measuring_2009}. Chicco and Jurman argue that the Matthews correlation coefficient often provides a more informative assessment than accuracy or F1 score in binary classification, particularly when class distributions are imbalanced \cite{chicco_advantages_2020}. For multiclass problems, Grandini et al.\ review a wider range of metric families and discuss their relative strengths and limitations \cite{grandini_metrics_2020}. More recent studies expand this discussion further: Rainio et al.\ examine evaluation metrics together with appropriate statistical testing procedures \cite{rainio_evaluation_2024}, while Miller reviews metric selection in genetics and genomics, illustrating the importance of aligning metric choice with domain-specific objectives and error structures \cite{miller_review_2024}.

In probabilistic classification, performance assessment extends beyond discrimination to include calibration. Guo et al.\ show that modern neural networks may achieve high predictive accuracy while remaining poorly calibrated and excessively confident in their predictions \cite{guo_calibration_2017}. This work highlights the importance of evaluating predicted probabilities directly, especially in applications where downstream decisions depend not only on the predicted class, but also on the reliability of the associated confidence estimates.

Regression evaluation has likewise generated a substantial and sometimes debated body of work. Early forecasting research examined alternative error measures for comparing predictive performance across different series and application contexts \cite{armstrong_error_1992,hyndman_another_2006}. Building on this tradition, Tofallis advocates improved measures of relative prediction accuracy for estimation and model selection \cite{tofallis_better_2015}, while Botchkarev proposes a broader conceptual framework for organizing regression error metrics in machine learning \cite{botchkarev_new_2019}. Chicco et al.\ argue that $R^2$ is often more interpretable than commonly used absolute or squared error measures \cite{chicco_coefficient_2021}, and Gao provides a more focused discussion of the practical meaning of explained variation \cite{gao_r-squared_2023}. At the same time, recent research emphasizes that scalar summary metrics alone may be insufficient. Verma proposes a more comprehensive approach in which residual analysis complements aggregate scores in order to reveal patterns that standard metrics may conceal \cite{verma_comprehensive_2025}.

A further strand of research addresses dependence structures in the data, particularly in spatial and spatio-temporal prediction settings. In such cases, standard random cross-validation may substantially overestimate model performance because observations that are close in space or time can leak information across training and test splits. Meyer et al.\ show that target-oriented validation strategies, such as leaving locations or time periods out, can lead to markedly different and more realistic estimates of predictive performance in spatio-temporal applications \cite{meyer_improving_2018}. They also demonstrate that spatially autocorrelated predictors can produce misleadingly optimistic results when variable selection and validation are not aligned with the intended prediction target \cite{meyer_importance_2019}. Similar conclusions appear in more recent application-specific studies. Sweet et al.\ report that cross-validation strategy affects both measured model skill and substantive interpretation in climate-agriculture research \cite{sweet_cross-validation_2023}. Koldasbayeva et al.\ propose principles for reducing bias in cross-validation for species distribution modeling \cite{koldasbayeva_foundation_2025}, while Radočaj et al.\ show that spatial cross-validation yields more conservative and realistic estimates than conventional folds in subfield maize yield prediction \cite{radocaj_comparative_2025}. Together, these studies demonstrate that evaluation design must reflect the dependency structure of the data and the expected deployment setting of the model.

Overall, the literature makes clear that model evaluation should not be viewed as a routine reporting step. Rather, it is a methodological challenge shaped by the interaction of resampling design, metric selection, calibration assessment, leakage prevention, and domain-specific data structure. Although prior work offers substantial guidance on each of these dimensions, it also consistently shows that no single metric or validation strategy is universally sufficient \cite{arlot_survey_2010,hand_measuring_2009,rainio_evaluation_2024}. The present paper builds on this understanding by offering an integrated discussion of evaluation principles, common pitfalls, and the practical considerations involved in selecting metrics and validation protocols that align with the true objective of a supervised learning task.

\section{Problem Statement}
\label{problemstatement}

The evaluation of supervised machine learning models is often treated as a routine stage of the modeling pipeline in which a familiar set of metrics is reported after training and testing. In practice, however, metric selection is itself a methodological problem. Evaluation metrics are not interchangeable summaries of predictive quality. Each metric imposes a different penalty structure on prediction errors, and these penalty structures interact differently with the statistical properties of the data and with the practical consequences of model failure. As a result, the same model may appear highly effective under one metric and inadequate under another, even when all metrics are computed from the same predictions.

This issue is especially important in data mining and knowledge discovery, where models are valued not only for statistical fit but also for their ability to support reliable decisions under realistic conditions. In binary classification, a model may achieve high Accuracy under severe class imbalance while performing poorly on the minority class. In multiclass classification, micro-averaged metrics may be dominated by the most prevalent classes and therefore mask weak performance on underrepresented categories. In regression, MAE and RMSE may lead to different conclusions depending on whether the application is more sensitive to average error or to extreme deviations. Likewise, a seemingly strong $R^2$ value may still coexist with structured residual patterns that reveal important local model weaknesses.

The central problem addressed in this paper is therefore not simply how to compute evaluation metrics, but how to determine which metrics provide reliable evidence of model utility under different data regimes. We treat model evaluation as a conditional assessment problem in which metric choice depends on three main factors: the type of prediction task, the structural properties of the dataset, and the application-specific cost of predictive error. Under this view, metric disagreement is not an anomaly or a reporting inconvenience. It is an expected consequence of applying metrics with different mathematical emphases to datasets with non-ideal statistical structure.

More specifically, the paper focuses on four broad conditions under which metric disagreement commonly arises in supervised learning. The first is class imbalance in binary classification, where metrics dominated by the majority class may overstate model quality. The second is asymmetric misclassification cost, where false negatives and false positives do not have equal practical consequences. The third is multiclass prevalence skew, where the choice between micro- and macro-averaging can substantially alter the interpretation of model performance. The fourth is heterogeneous regression error structure, where the relative usefulness of MAE, RMSE, and $R^2$ depends on outliers, heavy-tailed targets, and residual behavior.

These conditions are not exceptional cases. They are common characteristics of real-world machine learning problems. A major weakness of conventional evaluation practice is therefore that it often reports a standard set of metrics without first establishing whether those metrics are appropriate for the regime under study. This creates two related risks. First, model rankings may vary substantially across metrics, making comparison unstable. Second, and more importantly, deployment decisions may be based on evaluation criteria that are mathematically valid but practically misleading.

The present study addresses this problem through a scenario-driven empirical analysis designed to examine the relationship between dataset structure and metric reliability. Rather than treating benchmark datasets as interchangeable testbeds, we use them to represent distinct evaluation regimes that are likely to generate systematic metric disagreement. The objective is not only to compare algorithmic performance, but also to explain when and why commonly used metrics diverge, whether this divergence is stable across validation folds, and which metrics remain most informative under each condition.

Accordingly, this paper is guided by the following questions: under which dataset regimes do commonly reported metrics lead to different conclusions about model quality; which forms of disagreement can be explained by identifiable structural properties of the data; how should metric selection be aligned with the practical objective of the prediction task; and which metrics should be treated as primary decision criteria rather than secondary diagnostics.

The contribution of the paper is therefore methodological rather than purely descriptive. Evaluation is not treated as a fixed post hoc summary of predictions, but as a design choice that must be justified in relation to data characteristics and error costs. The experiments that follow are intended to support this perspective and to show that metric disagreement is a systematic and interpretable phenomenon rather than a statistical curiosity.

In summary, the problem addressed in this paper can be stated as follows: given a supervised learning task with known structural and operational constraints, how should evaluation metrics be selected and interpreted so that reported model performance remains aligned with relevant deployment notions of error? The remainder of the paper approaches this question through a structured experimental protocol and a set of representative case studies spanning both classification and regression.
\section{Evaluation metrics}
\label{evaluationmetrics}

Evaluating supervised machine learning models is essential for determining how well they generalize to unseen data. However, the choice of evaluation metric is not purely mathematical. It depends on the structure of the dataset, the nature of the prediction task, and the practical consequences of different types of errors. A metric that appears appropriate in one context may be misleading in another. For this reason, model evaluation must be aligned with both statistical properties and domain objectives. In this study, classification performance is assessed using Accuracy, Precision, Recall, F1 Score, Matthews Correlation Coefficient (MCC), ROC AUC, and PR AUC, while regression performance is evaluated using Mean Absolute Error (MAE), Root Mean Squared Error (RMSE), and the coefficient of determination ($R^2$). The equations presented below formalize these metrics and clarify the aspects of model behavior that each one emphasizes.

\subsection{Classification Evaluation Metrics}

Classification metrics are used to assess how well a model predicts categorical labels. Most of these measures are derived from the confusion matrix, which records four possible outcomes: True Positives ($TP$), True Negatives ($TN$), False Positives ($FP$), and False Negatives ($FN$). Because each metric is built from these quantities in a different way, each highlights a different aspect of classification performance.

\subsubsection{Accuracy}

Accuracy is the most direct classification metric and represents the proportion of correctly classified instances over the full dataset. As shown in Equation~\ref{eq:accuracy}, it is computed as the sum of true positives and true negatives divided by the total number of observations. Although Accuracy is simple and intuitive, it can be highly misleading in imbalanced datasets, where correct predictions on the majority class may dominate the score and conceal poor minority-class detection.

\begin{equation}
\label{eq:accuracy}
Accuracy = \frac{TP + TN}{TP + TN + FP + FN}
\end{equation}

\subsubsection{Precision and Recall}

Precision and Recall focus specifically on positive-class prediction, but they capture different priorities. As defined in Equation~\ref{eq:precision}, Precision measures the proportion of predicted positive instances that are actually positive. Recall, given in Equation~\ref{eq:recall}, measures the proportion of actual positive instances that are correctly identified by the model. In practice, these two metrics often involve a trade-off. Higher Precision reduces false positives, while higher Recall reduces false negatives. In applications such as medical diagnosis or anomaly detection, Recall is often more important because the cost of missing a true positive can be severe.

\begin{equation}
\label{eq:precision}
Precision = \frac{TP}{TP + FP}
\end{equation}

\begin{equation}
\label{eq:recall}
Recall = \frac{TP}{TP + FN}
\end{equation}

\subsubsection{F1 Score}

The F1 Score combines Precision and Recall into a single measure by taking their harmonic mean. As shown in Equation~\ref{eq:f1}, the harmonic formulation ensures that the score becomes low whenever either Precision or Recall is low. This makes the F1 Score especially useful in imbalanced settings where a model should not be rewarded for optimizing one of the two measures at the expense of the other.

\begin{equation}
\label{eq:f1}
F1 = 2 \cdot \frac{Precision \cdot Recall}{Precision + Recall}
\end{equation}

\subsubsection{Matthews Correlation Coefficient (MCC)}

The Matthews Correlation Coefficient provides a more balanced summary of binary classification performance by incorporating all four elements of the confusion matrix. As expressed in Equation~\ref{eq:mcc}, MCC accounts simultaneously for correct and incorrect predictions across both classes. For this reason, it is often regarded as a particularly informative metric when the data are imbalanced, since it avoids the overly optimistic interpretation that can arise from Accuracy alone.

\begin{equation}
\label{eq:mcc}
MCC = \frac{TP \cdot TN - FP \cdot FN}{\sqrt{(TP + FP)(TP + FN)(TN + FP)(TN + FN)}}
\end{equation}

\subsubsection{ROC AUC and PR AUC}

When classifiers produce probability scores rather than only class labels, threshold-independent metrics become particularly useful. The Receiver Operating Characteristic (ROC) curve plots the True Positive Rate against the False Positive Rate across all possible thresholds, and the Area Under this Curve is summarized by ROC AUC. As indicated in Equation~\ref{eq:rocauc}, ROC AUC evaluates how well the classifier ranks positive instances above negative ones across the full threshold range.

\begin{equation}
\label{eq:rocauc}
ROC \, AUC = \int_{0}^{1} TPR(FPR)\, d(FPR)
\end{equation}

Although ROC AUC is widely used, it can be deceptive under severe class imbalance because the False Positive Rate is normalized by the total number of actual negatives. In such cases, substantial numbers of false positives may still result in a low False Positive Rate. For this reason, the Precision--Recall curve is often more informative in skewed domains. As shown in Equation~\ref{eq:prauc}, PR AUC summarizes the relationship between Precision and Recall across thresholds, focusing more directly on minority-class performance.

\begin{equation}
\label{eq:prauc}
PR \, AUC = \int_{0}^{1} Precision(Recall)\, d(Recall)
\end{equation}

\subsection{Regression Evaluation Metrics}

Unlike classification, regression models predict continuous numerical values. Their evaluation therefore depends on measuring how far predicted values deviate from the true target values. Different regression metrics emphasize different error characteristics, such as average magnitude, sensitivity to outliers, or proportion of explained variation.

\subsubsection{Mean Absolute Error (MAE)}

Mean Absolute Error measures the average absolute distance between the predicted values and the true observations. As defined in Equation~\ref{eq:mae}, MAE computes the mean of the absolute residuals, giving equal linear weight to all errors regardless of direction. Because of this, MAE is generally easy to interpret and relatively robust to extreme outliers.

\begin{equation}
\label{eq:mae}
MAE = \frac{1}{n}\sum_{i=1}^{n} |y_i - \hat{y}_i|
\end{equation}

\subsubsection{Root Mean Squared Error (RMSE)}

Root Mean Squared Error measures the square root of the average squared residuals. As shown in Equation~\ref{eq:rmse}, the squaring operation causes larger errors to contribute disproportionately more to the final score. This makes RMSE more sensitive to outliers than MAE and particularly useful when large prediction errors are especially undesirable.

\begin{equation}
\label{eq:rmse}
RMSE = \sqrt{\frac{1}{n}\sum_{i=1}^{n}(y_i - \hat{y}_i)^2}
\end{equation}

\subsubsection{$R^2$ (Coefficient of Determination)}

The coefficient of determination, $R^2$, evaluates the proportion of variance in the target variable that is explained by the model relative to a baseline predictor based on the sample mean. As formalized in Equation~\ref{eq:r2}, it compares the residual sum of squares to the total sum of squares. A higher $R^2$ indicates that the model explains a larger share of the observed variation. However, it does not directly express the magnitude of prediction error, so it should be interpreted alongside error-based measures such as MAE and RMSE.

\begin{equation}
\label{eq:r2}
R^2 = 1 - \frac{\sum_{i=1}^{n}(y_i - \hat{y}_i)^2}{\sum_{i=1}^{n}(y_i - \bar{y})^2}
\end{equation}

\section{Datasets}
\label{datasets}

To provide a broad and credible experimental basis for the evaluation of supervised machine learning models, this study uses 15 real-world datasets obtained from established public repositories, including the UCI Machine Learning Repository and OpenML. The selected datasets were chosen to reflect a wide range of structural properties and application domains, rather than relying on overly simplified benchmark problems. In particular, they include class imbalance, multiclass structure, sparse high-dimensional text representations, categorical decision spaces, large-scale observations, and continuous targets with different noise and outlier characteristics. This diversity is essential for examining how evaluation metrics behave under realistic conditions and for revealing situations in which a single metric may produce an incomplete or misleading impression of performance.

\subsection{Classification Datasets}

For the classification experiments, 10 datasets were selected to cover a range of binary and multiclass prediction settings. These datasets were chosen to reflect several important evaluation challenges, including severe class imbalance, asymmetric misclassification costs, categorical and text-based input spaces, and large differences in the number of classes. As shown in Table~\ref{tab:classification_datasets}, the classification benchmark includes domains such as finance, medicine, agriculture, ecology, text analysis, image recognition, business analytics, demographics, and automotive decision support. The variation in dataset size, dimensionality, and class structure makes this collection suitable for comparing metrics such as Accuracy, F1 Score, MCC, ROC AUC, and PR AUC under substantially different conditions.

The Default of Credit Card Clients dataset is included because it represents a financial binary classification problem with a skewed class distribution, making it well suited for demonstrating how Accuracy may overstate practical model effectiveness \cite{yeh_comparisons_2009}. The Dry Bean dataset provides a multiclass setting with seven categories, allowing comparison between alternative averaging schemes such as macro- and micro-averaged F1 \cite{koklu_multiclass_2020}. The Chronic Kidney Disease dataset represents a medical diagnostic setting in which false negatives are especially serious, making it appropriate for examining the importance of Recall in high-risk applications \cite{rubini_chronic_2015}. The Letter Recognition dataset introduces a larger multiclass image-derived problem with 26 output classes, while the 20 Newsgroups dataset contributes a sparse and high-dimensional text classification setting that is useful for observing evaluation behavior in natural language processing tasks \cite{frey_letter_1991,lang_newsweeder_1995}.

The Adult Census Income dataset was selected because it supports the study of threshold-dependent classification, calibration, and cost-sensitive decision making in a demographic prediction context \cite{kohavi_scaling_1996}. The Covertype dataset adds a much larger multiclass problem, providing a demanding setting for volume-sensitive evaluation and large-scale predictive performance analysis \cite{blackard_comparative_1999}. The Bank Marketing dataset reflects a business conversion problem in which the practical value of a classifier depends on how well the selected metric aligns with decision-making objectives \cite{moro_data_2014}. The Breast Cancer Wisconsin (Diagnostic) dataset provides a high-dimensional medical binary classification problem that is useful for discussing asymmetric decision costs in clinical settings \cite{wolberg_breast_1995}. Finally, the Car Evaluation dataset contributes a purely categorical decision problem, enabling the analysis of metric behavior in the absence of continuous feature boundaries \cite{bohanec_knowledge_1988}.

As can be seen in Table~\ref{tab:classification_datasets}, the classification datasets vary substantially in scale, from a few hundred observations to several hundred thousand, and range from binary tasks to 26-class recognition problems. This heterogeneity is useful because it allows the experimental evaluation to move beyond a single type of classification setting and instead test metric behavior under a range of realistic data conditions.

\begin{table}[t]
\caption{Summary of classification datasets}
\label{tab:classification_datasets}
\centering
\begin{tabular}{llllrr}
\hline
ID & Dataset & Domain & Instances & Features & Classes \\
\hline
A & Default of Credit Card & Finance & 30000 & 23 & 2 \\
B & Bank Marketing & Business & 45211 & 16 & 2 \\
C & Breast Cancer Diagnostic & Medical & 569 & 30 & 2 \\
D & Chronic Kidney Disease & Medical & 400 & 24 & 2 \\
E & Dry Bean & Agriculture & 13611 & 16 & 7 \\
F & Covertype & Ecology & 581012 & 54 & 7 \\
G & 20 Newsgroups & Text & 18846 & Sparse & 20 \\
H & Letter Recognition & Image & 20000 & 16 & 26 \\
I & Adult Census Income & Demographics & 48842 & 14 & 2 \\
J & Car Evaluation & Automotive & 1728 & 6 & 4 \\
\hline
\end{tabular}
\end{table}

\subsection{Regression Datasets}

For the regression experiments, five datasets were selected to capture different continuous prediction challenges, including heavy-tailed targets, outlier sensitivity, biomedical prediction, count-related demand variation, and relatively clean baseline regression structure. These datasets support the comparative analysis of MAE, RMSE, and $R^2$, each of which responds differently to scale, variance, and extreme deviations. Table~\ref{tab:regression_datasets} summarizes the main characteristics of the regression datasets used in this study.

The California Housing dataset was selected because it represents a real-world spatial prediction problem in which the target variable exhibits substantial heterogeneity, making it useful for examining the limitations of scalar summary metrics on uneven target distributions \cite{pace_sparse_1997}. The Individual Household Electric Power Consumption dataset introduces a much larger time-series setting with more than two million measurements, making it particularly suitable for comparing the robustness of MAE against the stronger outlier sensitivity of RMSE during periods of extreme consumption \cite{hebrail_individual_2012}. The Parkinson Telemonitoring dataset provides a medical regression problem in which biomedical voice features are used to predict clinician-assessed UPDRS scores \cite{tsanas_accurate_2009}. The Seoul Bike Sharing Demand dataset contributes a demand forecasting problem influenced by seasonal and environmental variation, offering a useful example of count-oriented prediction with continuous evaluation criteria \cite{seoul_bike_sharing_2020}. The Diabetes Progression dataset serves as a relatively clean medical regression benchmark and is particularly appropriate for baseline model comparison and residual analysis \cite{efron_least_2004}.

As shown in Table~\ref{tab:regression_datasets}, the regression datasets also display considerable variation in size and application domain. This breadth is important because it allows the study to compare regression metrics not only under idealized conditions, but also in settings involving large sample sizes, medical outcomes, transportation demand, and real-estate valuation. Such variation helps illustrate why no single regression metric is universally sufficient and why metric choice must be matched to the structure and practical meaning of the target variable.

\begin{table}[t]
\caption{Summary of regression datasets}
\label{tab:regression_datasets}
\centering
\begin{tabular}{llllrl}
\hline
ID & Dataset & Domain & Instances & Features & Target type \\
\hline
K & California Housing & Real Estate & 20640 & 8 & Continuous \\
L & Power Consumption & Energy & 2075259 & 9 & Continuous \\
M & Parkinson Telemonitoring & Medical & 5875 & 19 & Continuous \\
N & Seoul Bike Sharing & Transportation & 8760 & 13 & Count data \\
O & Diabetes Progression & Medical & 442 & 10 & Continuous \\
\hline
\end{tabular}
\end{table}

\section{Experimental Evaluation}
\label{evaluation}

\subsection{Cross-Validation versus Hold-Out Approaches}

To strengthen the reliability of the empirical analysis, the evaluation framework was designed to move beyond a single 80:20 hold-out split and instead adopt a 5-fold cross-validation procedure. A hold-out approach is computationally simple and widely used, but it is also highly sensitive to the particular partition selected. When the training and test subsets are created only once, the reported performance can depend heavily on whether the split happens to be favorable or unfavorable. As a result, a single hold-out estimate may reflect sampling luck as much as genuine model quality.

To reduce this source of variability, we employ 5-fold cross-validation throughout the experimental study although it is common to use a 10-fold approach as well. Under this procedure, each dataset is divided into five mutually exclusive subsets of approximately equal size. The model is trained on four folds and evaluated on the remaining fold, and this process is repeated five times so that each observation serves as test data exactly once. The final performance estimate is then obtained by aggregating the results across all folds. This approach produces a more stable estimate of generalization performance and makes it possible to observe whether a model performs consistently or fluctuates substantially across different partitions of the data.

For classification problems, stratified folds are used in order to preserve the original class distribution within each split. This is especially important for imbalanced datasets, where random partitioning without stratification may distort the minority-to-majority ratio and produce misleading results. In addition, the evaluation is not tied to a single manually fixed partition, allowing the models to be assessed under realistic variation in the data composition. Reporting both fold-level behavior and average performance therefore provides a more robust basis for comparing metrics and algorithms across datasets with different structural properties.

\subsection{Classification Results}

The classification experiments reveal that metric disagreement is not an occasional anomaly, but a recurring and structurally important feature of supervised learning evaluation. Across the 10 classification datasets, models often appeared strong under one metric while showing clear weaknesses under another. To examine this phenomenon in a systematic way, the analysis is organized around four representative scenarios: imbalanced binary classification, asymmetrical misclassification costs in medical diagnosis, multiclass averaging strategies, and probability calibration under categorical decision spaces.

\subsubsection{Scenario 1: The Accuracy Trap in Highly Imbalanced Distributions}

The first scenario examines two imbalanced binary datasets, Default of Credit Card Clients and Bank Marketing, in order to demonstrate how traditional metrics can overstate model effectiveness when the negative class dominates the distribution. In both cases, the class imbalance creates a situation in which a model can achieve high overall Accuracy even if it performs inadequately on the minority class. This makes the scenario particularly suitable for comparing metrics that respond differently to imbalance, such as Accuracy, ROC AUC, PR AUC, and MCC.

The cross-validation results show that both datasets produce strong-looking scores under conventional global metrics. On the Default of Credit Card Clients dataset, the Random Forest model achieved an average Accuracy of 0.8150 and an average ROC AUC of 0.7565 across the five folds. On the Bank Marketing dataset, the same model achieved an even higher average Accuracy of 0.9031 together with an average ROC AUC of 0.9158. If these metrics were viewed in isolation, one might conclude that the classifiers are highly effective and suitable for practical deployment.

However, the interpretation changes substantially when attention shifts to imbalance-sensitive metrics. On the Credit Card dataset, the average PR AUC fell to 0.5230 and the average MCC to 0.3851, indicating that the classifier’s ability to identify the minority class was much weaker than suggested by Accuracy alone. A similar pattern appears in the Bank Marketing dataset, where an average ROC AUC of 0.9158 coincided with a PR AUC of only 0.5855 and an MCC of 0.4460. These results show that the apparently strong ranking and classification performance is partly an artifact of the large number of true negatives.

\begin{figure}
    \centering
    \includegraphics[width=1.0\linewidth]{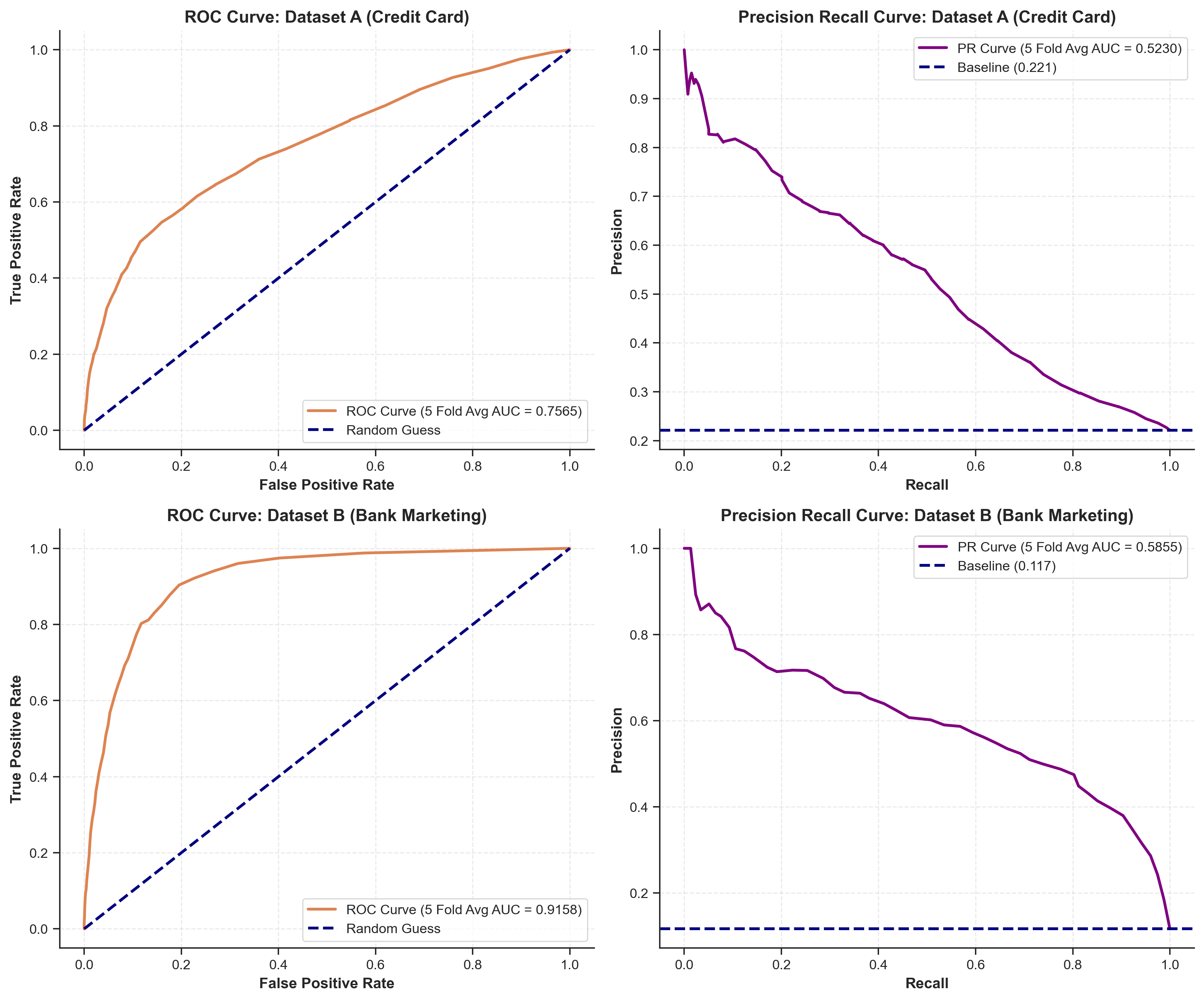}
    \caption{Comparison of evaluation behavior on two imbalanced classification datasets.}
    \label{fig1}
\end{figure}

This contrast is illustrated clearly in Figure~\ref{fig1}, where the performance patterns of the two imbalanced datasets are presented together. As shown in Figure~\ref{fig1}, Accuracy and ROC AUC remain comparatively high, whereas PR AUC and MCC expose the underlying limitations of the models on the minority class. The figure therefore supports the argument that in highly imbalanced financial and business problems, evaluation should not be based primarily on Accuracy. Instead, metrics that are more sensitive to minority-class performance provide a more honest basis for judging whether a model is operationally useful.

In practical terms, this scenario shows that metric choice must reflect the real business cost of errors. In credit risk modeling, missing a likely default may create direct financial loss, while in marketing, failing to identify a potential conversion target may reduce campaign effectiveness. For such settings, PR AUC and MCC provide a more realistic assessment than Accuracy alone because they are less easily inflated by the dominant majority class.

\subsubsection{Scenario 2: Asymmetrical Misclassification Costs in Diagnostics}

The second scenario focuses on medical diagnostics, where the consequences of false negatives and false positives are fundamentally unequal. Using the Breast Cancer Wisconsin Diagnostic dataset and the Chronic Kidney Disease dataset, the experiments investigate how standard classification metrics may fail to reflect the true safety requirements of clinical decision support.

The Chronic Kidney Disease dataset produced nearly perfect results under Random Forest classification, including an average Recall of 1.0000 and an average Accuracy of 0.9950. This suggests that the class boundary in that dataset is relatively easy for the model to learn. The Breast Cancer dataset, however, provides a more realistic and challenging diagnostic scenario. On this dataset, the same model achieved an average Accuracy of 0.9648 and an average Precision of 0.9580, values that might initially appear strong enough to justify deployment.

A closer inspection of Recall reveals a more important limitation. The average Recall was 0.9481, and in one validation fold it dropped to 0.9286. In a clinical context, this means that the model failed to identify a non-trivial proportion of malignant cases. Although the F1 Score averaged 0.9526, that result partly masks the severity of the issue because F1 gives equal weight to Precision and Recall. In settings where missing a positive diagnosis is substantially more costly than producing a false alarm, a recall-oriented measure such as the F2 Score provides a more appropriate perspective. In this scenario, the average F2 Score fell to 0.9498, reflecting more clearly the clinical significance of missed detections.

\begin{figure}
    \centering
    \includegraphics[width=1.0\linewidth]{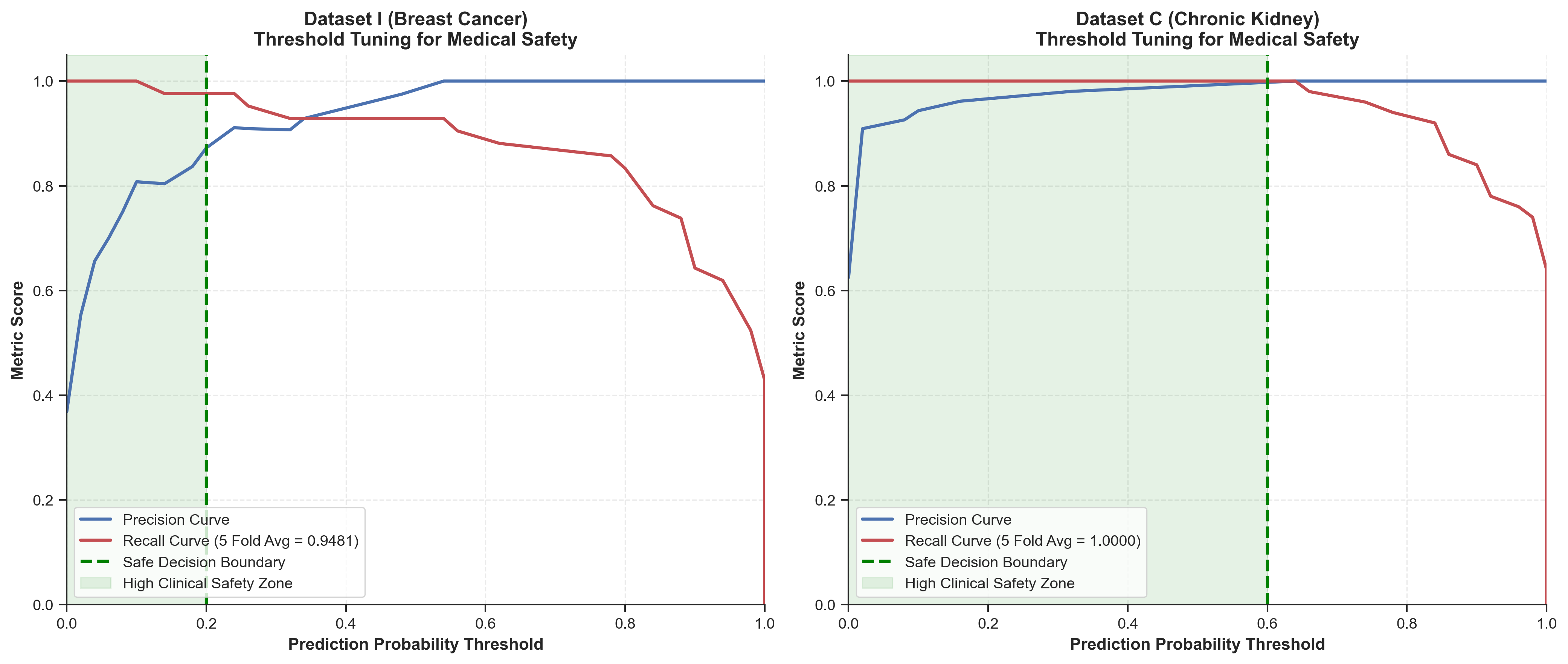}
    \caption{Threshold trade-offs in medical classification, emphasizing the interaction between Precision and Recall.}
    \label{fig2}
\end{figure}

This trade-off is visualized in Figure~\ref{fig2}. As shown in Figure~\ref{fig2}, increasing Precision tends to suppress Recall, and vice versa, which makes default decision thresholds unsuitable for safety-critical medical applications. The figure highlights the need for threshold tuning when the evaluation objective is not merely balanced performance, but the reduction of false negatives to the lowest possible level.

Taken together, the results of this scenario show that clinical model evaluation cannot rely on Accuracy or even F1 Score alone. When the cost of a false negative is potentially severe, the evaluation framework must prioritize Recall and related recall-weighted metrics, while also using decision threshold tuning to align classifier behavior with the safety requirements of the medical domain.

\subsubsection{Scenario 3: Averaging Strategies in Multiclass Distributions}

The third scenario considers multiclass classification and investigates the difference between micro- and macro-averaged performance. Four datasets were selected for this purpose: Dry Bean, Covertype, 20 Newsgroups, and Letter Recognition. Together, these datasets represent a diverse set of multiclass challenges, including sparse high-dimensional text data, ecological class imbalance, and relatively uniform class structure.

The results reveal a consistent mathematical relationship between Accuracy and Micro F1. Across all folds and all four datasets, the Micro F1 score matched global Accuracy, with average values of 0.9226 for Dry Bean, 0.8916 for Covertype, 0.6879 for 20 Newsgroups, and 0.9615 for Letter Recognition. This is expected because Micro F1 aggregates decisions across all classes before computing the score and is therefore dominated by the largest classes, in much the same way as Accuracy.

The difference between Micro F1 and Macro F1 becomes most informative when class balance is uneven. In Covertype, the average Micro F1 was 0.8916, but the average Macro F1 dropped to 0.8425, indicating that the classifier performed substantially worse on minority forest cover types than on the dominant classes. By contrast, the Dry Bean dataset showed the reverse pattern, with Macro F1 slightly exceeding Micro F1, suggesting that the classifier handled the smaller classes comparatively well. Letter Recognition served as a near-control case: because the class distribution is close to uniform, the two metrics converged almost exactly, with averages of 0.9615 and 0.9614.

\begin{figure}
    \centering
    \includegraphics[width=1.0\linewidth]{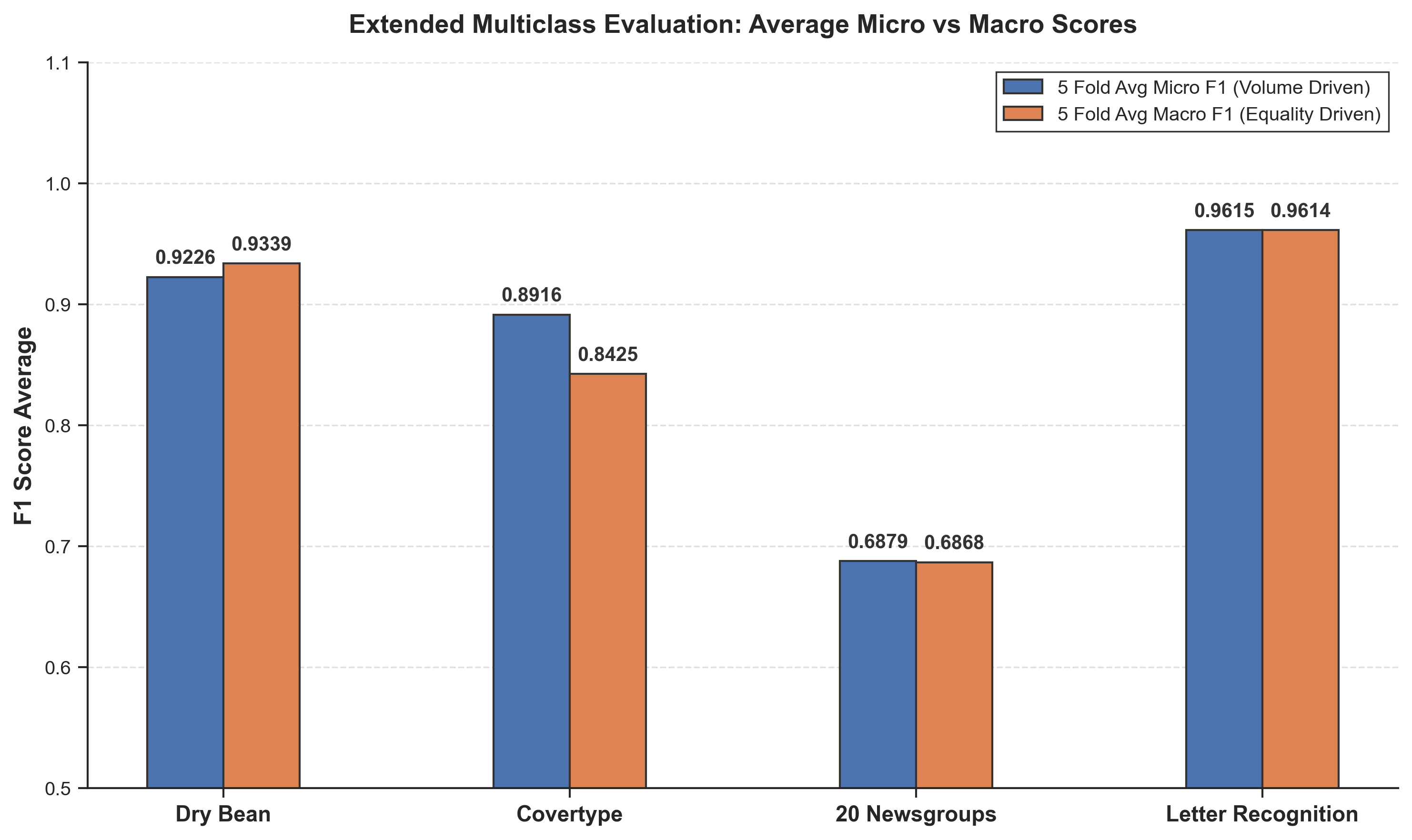}
    \caption{Comparison of Micro F1 and Macro F1 across multiclass datasets with different class distributions.}
    \label{fig3}
\end{figure}

These patterns are summarized in Figure~\ref{fig3}. As shown in Figure~\ref{fig3}, divergence between Micro F1 and Macro F1 becomes more pronounced when the underlying class distribution is uneven, whereas the two measures converge in more balanced multiclass settings. The figure therefore makes clear that the choice between averaging strategies is not a technical detail, but a substantive decision about what aspect of performance should be emphasized.

This scenario shows that metric selection in multiclass problems must be linked directly to the application objective. If the priority is maximizing overall throughput across all instances, Micro F1 is appropriate. If the priority is more equitable performance across all classes, particularly the rare ones, Macro F1 is the more informative choice.

\subsubsection{Scenario 4: Probability Calibration and the Log Loss Penalty}

The fourth classification scenario addresses probability calibration. Two datasets with strong categorical structure, Adult Census Income and Car Evaluation, were used to compare Decision Trees and Random Forests under both Accuracy and Log Loss. The purpose of this analysis is to show that classification correctness alone may conceal serious weaknesses in predicted probabilities.

On the Car Evaluation dataset, the Decision Tree slightly outperformed the Random Forest in average Accuracy, scoring 0.9167 compared with 0.9039. If model selection were based on Accuracy alone, the Decision Tree would appear preferable. However, the comparison changes completely when Log Loss is considered. The Random Forest achieved an average Log Loss of 0.3350, while the Decision Tree produced a much worse average value of 3.0035. This indicates that although the tree often made correct decisions, its incorrect predictions were delivered with extreme overconfidence.

A similar pattern appears in the Adult Census dataset. The Accuracy values of the two models were relatively close, but the Decision Tree again produced a substantially worse Log Loss than the Random Forest. This confirms that uncalibrated classifiers can appear competitive under discrete accuracy-based metrics while being unreliable in probabilistic terms.

\begin{figure}
    \centering
    \includegraphics[width=1.0\linewidth]{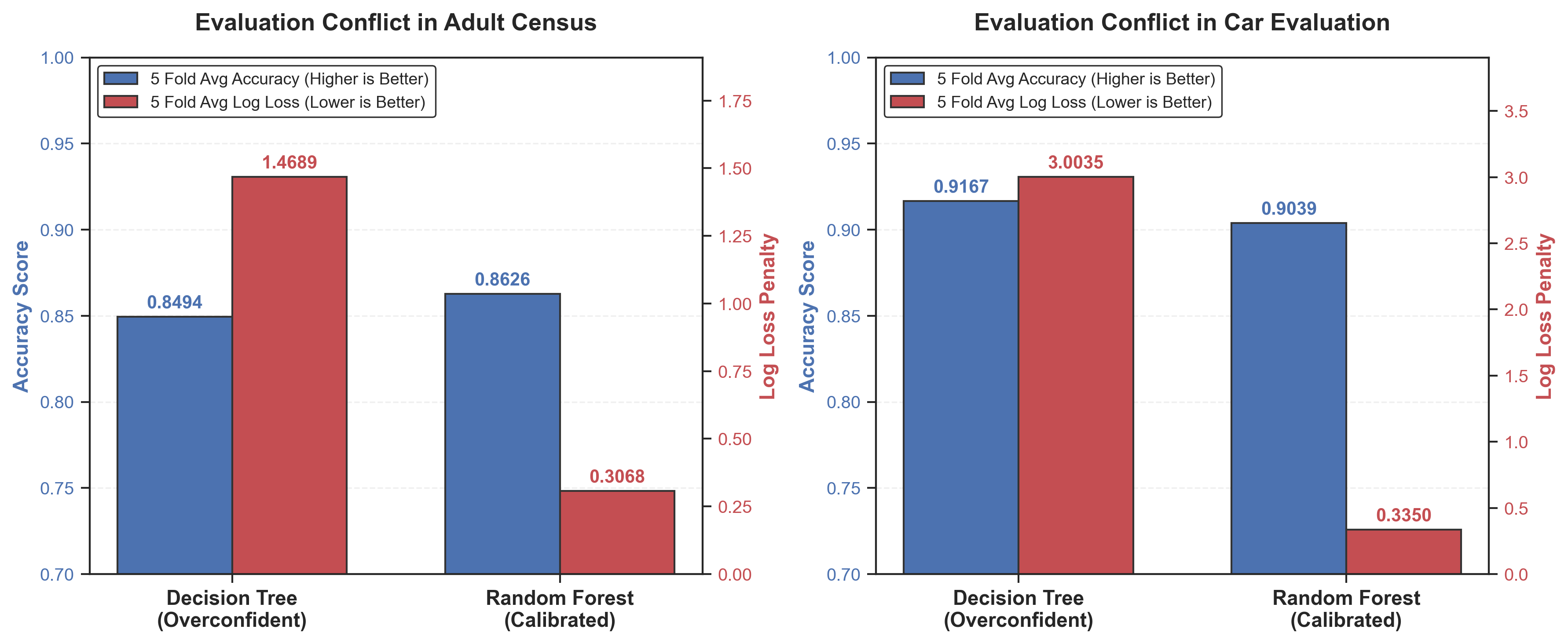}
    \caption{Contrast between Accuracy and Log Loss, showing how calibration can alter model ranking.}
    \label{fig4}
\end{figure}

Figure~\ref{fig4} illustrates this inversion in model ranking. As shown in Figure~\ref{fig4}, the model that appears better under Accuracy is not necessarily the one that performs better under Log Loss. The figure highlights the importance of calibration-sensitive evaluation whenever predicted probabilities are intended to support downstream decisions.

This scenario demonstrates that model evaluation in risk-sensitive applications should not rely only on whether class labels are correct. When decisions depend on predicted probabilities, Log Loss becomes essential because it penalizes unwarranted confidence and rewards better-calibrated uncertainty estimates.

\subsection{Regression Results}

The regression experiments reveal that different error metrics capture distinct aspects of predictive quality. Across the five regression datasets, the results show that the interpretation of model performance depends strongly on whether the evaluation emphasizes average absolute error, sensitivity to large deviations, explained variance, or relative error behavior near low baselines. To make these differences explicit, the regression analysis is organized into three scenarios: outlier sensitivity, the interpretation of $R^2$ with residual analysis, and the instability of percentage-based errors.

\subsubsection{Scenario 5: The Outlier Penalty in Regression Error Metrics}

The first regression scenario compares MAE and RMSE on two datasets with heavy-tailed behavior: California Housing and Individual Household Electric Power Consumption. These datasets were chosen because both contain rare but important extreme values. In California Housing, luxury properties create unusually large target values, while in the electricity dataset, short-lived consumption spikes introduce abrupt deviations from normal behavior.

The results show a clear divergence between MAE and RMSE. On the California Housing dataset, the Random Forest model achieved an average MAE of 0.3308, whereas the average RMSE rose to 0.5085. This indicates that the model performed reasonably well on typical observations but incurred substantially larger penalties on the most extreme house values. The same effect was even stronger in the Power Consumption dataset, where the model achieved an average MAE of 0.0216 but an average RMSE of 0.0401. In one fold, the RMSE inflation became especially large, reflecting the presence of sharp power spikes that had a disproportionate influence on the squared-error metric.

\begin{figure}
    \centering
    \includegraphics[width=1.0\linewidth]{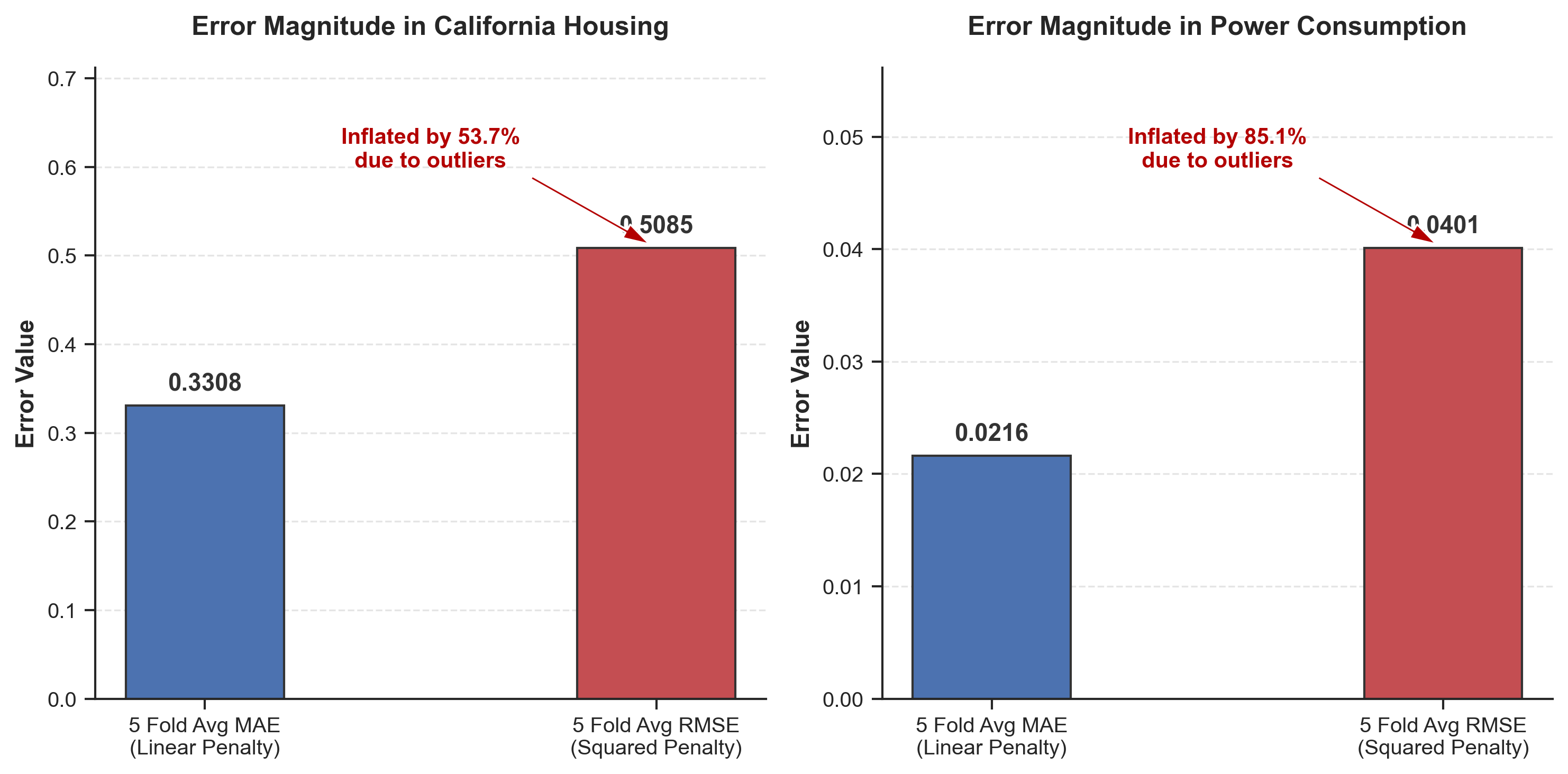}
    \caption{Comparison of MAE and RMSE under heavy-tailed regression errors.}
    \label{fig5}
\end{figure}

This divergence is illustrated in Figure~\ref{fig5}. As shown in Figure~\ref{fig5}, RMSE consistently exceeds MAE by a substantial margin in both datasets, which reflects the stronger penalty assigned to large residuals. The figure therefore clarifies why RMSE is more sensitive to outliers and why MAE can provide a more robust description of central predictive behavior.

This scenario shows that regression metric selection must be guided by the meaning of large errors in the application domain. If extreme deviations are rare anomalies that should not dominate the evaluation, MAE is more appropriate. If large errors represent operationally dangerous failures, RMSE provides a more suitable penalty structure.

\subsubsection{Scenario 6: The Illusion of $R^2$ and the Necessity of Residual Analysis}

The second regression scenario examines the limitations of relying on $R^2$ as a standalone indicator of model quality. The Diabetes Progression and Parkinson Telemonitoring datasets were used here because they represent medical regression tasks with different levels of predictability and noise.

On the Parkinson dataset, the Random Forest model achieved an average $R^2$ of 0.9673 and an average MAE of 0.8398, suggesting excellent predictive performance. On the Diabetes dataset, by contrast, the average $R^2$ was much lower at 0.4023 and varied across folds. If the analysis stopped at these scalar summaries, one might conclude that the Parkinson model is nearly flawless and that the Diabetes model is comparatively weak.

However, scalar measures alone do not show whether errors are randomly distributed or whether the model behaves systematically worse in certain regions of the target space. Residual analysis is therefore necessary. Plotting residuals against predicted values allows the detection of patterns such as curvature, heteroscedasticity, or systematic underprediction at higher target levels. In the Parkinson dataset, such analysis is particularly important to verify that the model does not become less reliable for patients with more severe symptoms. In the Diabetes dataset, residual structure can help determine whether the modest $R^2$ reflects genuine model inadequacy or simply the inherent variability of the clinical outcome.

\begin{figure}
    \centering
    \includegraphics[width=1.0\linewidth]{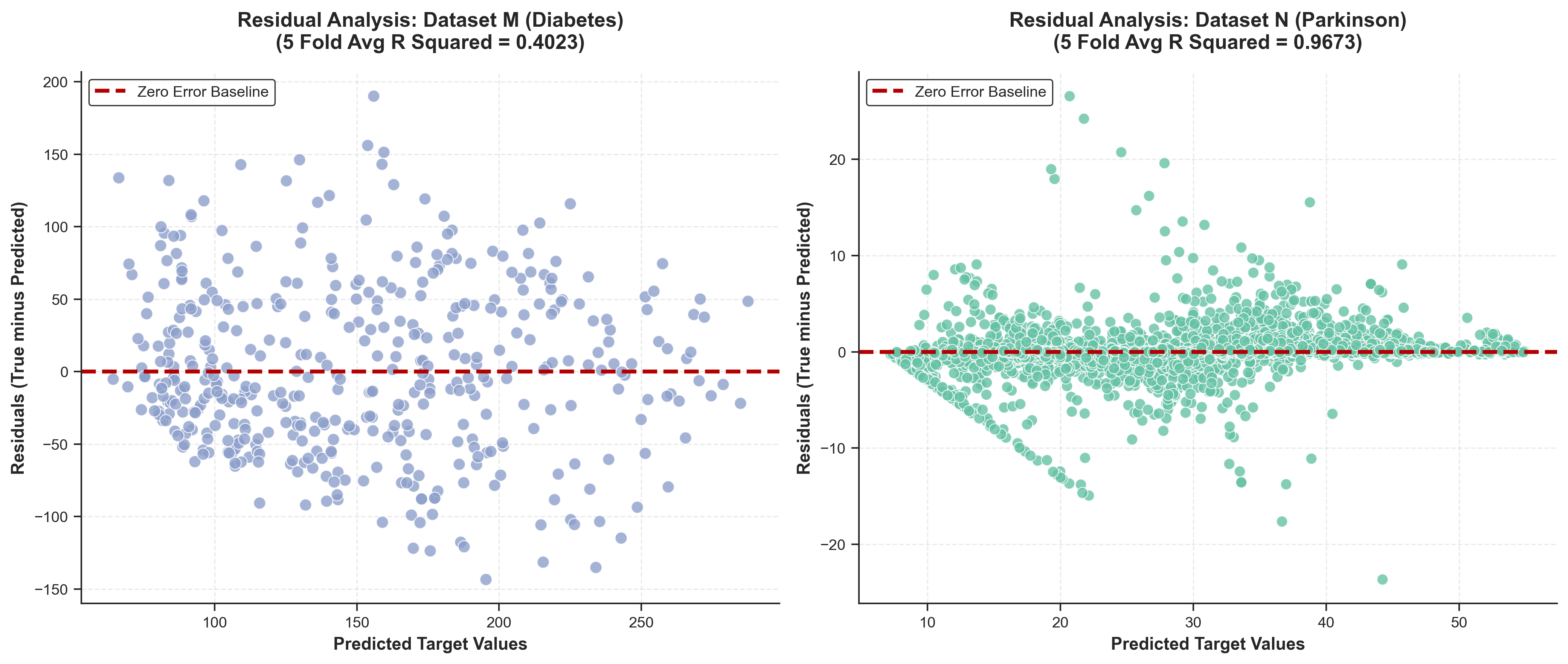}
    \caption{Residual analysis illustrating why global goodness-of-fit measures should be complemented by error diagnostics.}
    \label{fig6}
\end{figure}

This point is illustrated in Figure~\ref{fig6}. As shown in Figure~\ref{fig6}, residual plots provide information that is not available from $R^2$ alone, making it possible to distinguish between random noise and structured prediction bias. The figure therefore supports the broader conclusion that regression evaluation should not be based solely on global fit measures.

In high-stakes settings such as medical prediction, this scenario shows that even a very high $R^2$ can create a false sense of security if residual structure is ignored. Proper evaluation therefore requires both scalar metrics and diagnostic visual analysis.

\subsubsection{Scenario 7: Baseline Sensitivity and the Asymmetry of Percentage Errors}

The final regression scenario investigates the instability of percentage-based evaluation when the target variable can take very small values. The Seoul Bike Sharing Demand and Power Consumption datasets were used because both contain intervals with low or near-zero demand, making them suitable for illustrating the weaknesses of Mean Absolute Percentage Error (MAPE).

The results show that relatively small absolute deviations can translate into inflated percentage errors when the baseline is close to zero. On the Power Consumption dataset, the model achieved an average MAE of 0.0216, indicating very accurate prediction in absolute terms. Nevertheless, the average MAPE was 3.4872\%, reflecting the fact that small denominators can magnify minor deviations. A similar effect was observed in the Seoul Bike Sharing dataset, where the average MAE was 14.5654 but the average MAPE reached 16.6201\%. In operational terms, an absolute error of about 14 bicycles may be minor during busy periods, yet the percentage-based metric becomes dominated by low-demand intervals.

\begin{figure}
    \centering
    \includegraphics[width=1.0\linewidth]{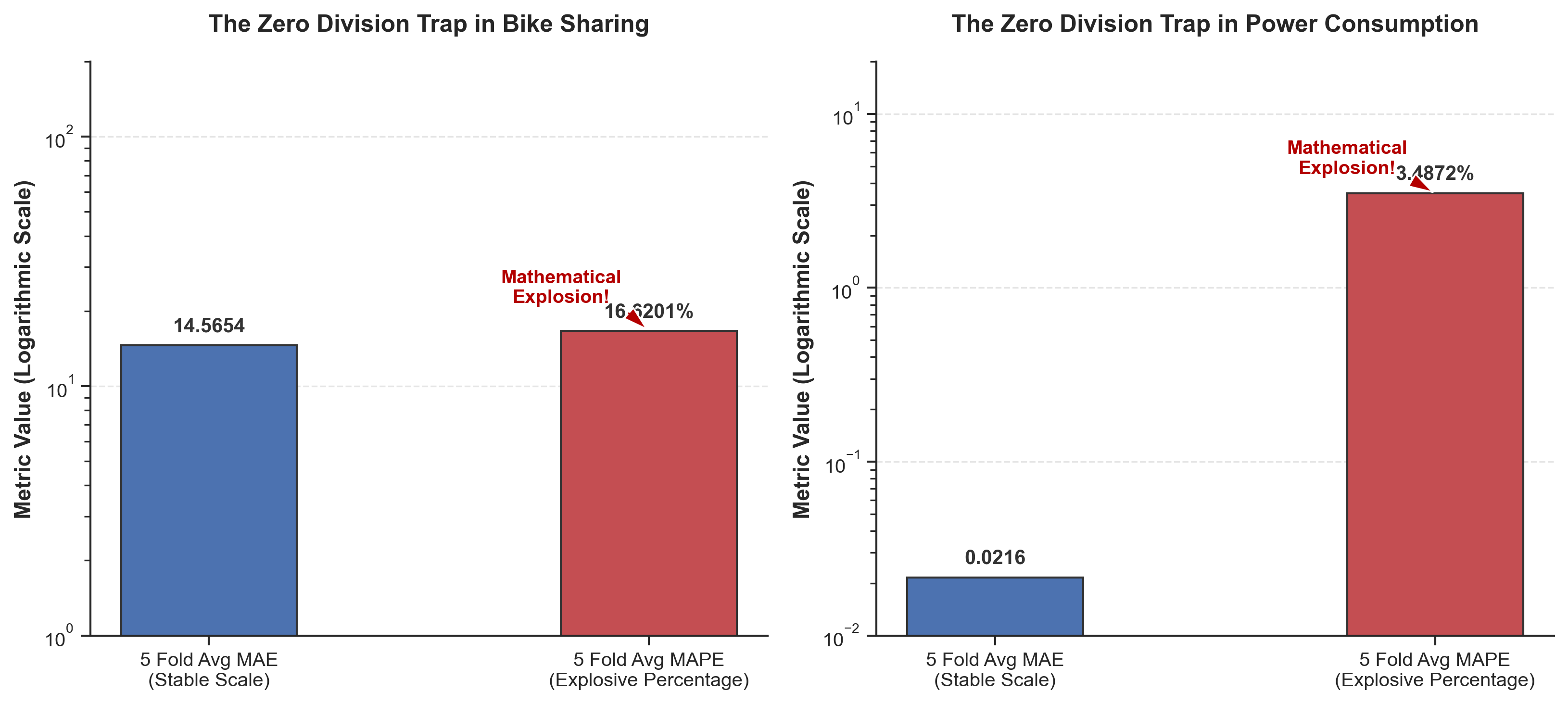}
    \caption{Illustration of the instability of percentage-based error metrics near low baseline values.}
    \label{fig7}
\end{figure}

This instability is illustrated in Figure~\ref{fig7}. As shown in Figure~\ref{fig7}, percentage errors can escalate sharply when the target approaches zero, even when the underlying absolute prediction errors remain modest. The figure highlights why MAPE can be misleading in intermittent-demand or low-baseline forecasting problems.

This scenario demonstrates that MAPE is not universally appropriate for regression evaluation. In applications with frequent low-demand or near-zero target values, absolute metrics such as MAE provide a more stable assessment, while alternative relative metrics should be considered only when their denominator behavior is well aligned with the structure of the data.

\subsection{Discussion}

A consistent pattern across the experimental results presented in this section is the phenomenon of metric disagreement. In multiple scenarios, models that appear highly effective under one evaluation criterion perform poorly when assessed using another. For example, in Scenario 1, a Random Forest achieved more than 81\% Accuracy while obtaining only 0.37 MCC, indicating that the apparently strong overall classification performance concealed important weaknesses in balanced predictive quality. Likewise, in Scenario 6, Decision Trees produced comparatively strong Accuracy values but performed very poorly under Log Loss, revealing severe overconfidence and poor probabilistic calibration.

This disagreement between metrics should not be interpreted as a mathematical inconsistency. Rather, it reflects a fundamental property of model evaluation. Each metric captures a different dimension of predictive behavior and imposes its own implicit assumptions about what types of errors matter most. Some metrics reward overall correctness, others emphasize minority class detection, while others penalize overconfident or disproportionately large errors. As soon as the dataset departs from an idealized setting of balanced classes, symmetric costs, and stable distributions, these different metric perspectives naturally begin to diverge. What appears to be good performance under one definition may represent poor performance under another.

For this reason, resolving metric disagreement requires moving away from a purely score-driven perspective and adopting a domain-centered interpretation of model quality. The goal is not to choose the metric that gives the most favorable numerical result, but to identify the one that most faithfully reflects the real consequences of prediction errors in the target application. In malicious traffic classification, for instance, a false negative may allow an attack to pass undetected and therefore impose far greater damage than a false positive, which may only trigger an unnecessary alert. In such settings, metrics such as Recall and PR AUC provide a more meaningful basis for evaluation than overall Accuracy or even ROC AUC, which may appear strong while obscuring poor minority-class detection. The same principle applies across other domains, including medical diagnosis, financial risk prediction, and business decision support, where the severity and asymmetry of errors must directly shape evaluation priorities.

The empirical findings of this study therefore support several practical lessons for supervised learning evaluation. First, Accuracy should not be treated as the default metric, especially when class imbalance is present. In these cases, more informative alternatives such as MCC or PR AUC often provide a more reliable picture of model behavior. Second, in multiclass classification, macro-averaged metrics are often preferable when the objective is to evaluate performance more equitably across all classes, including underrepresented ones, whereas micro-averaged metrics may be more suitable only when the main concern is overall instance-level performance. Third, regression metrics must be selected in accordance with the actual penalty structure of the problem. MAE is often preferable when robust and interpretable average error reporting is needed, while RMSE is more appropriate when large deviations are especially costly and should be penalized more heavily. Finally, scalar metrics alone are rarely sufficient. Measures such as $R^2$ should be complemented by residual analysis, and classification models that output probabilities should also be examined through calibration-sensitive measures such as Log Loss in order to detect overconfidence and hidden structural bias.

Taken together, these observations reinforce the broader argument of this paper: model evaluation is not a matter of reporting whichever metric appears most favorable, but of constructing an assessment framework that reflects the statistical properties of the data and the real-world objective of the task. A careful evaluation process must therefore be explicit about what kind of errors matter, how they are penalized, and whether the chosen metrics genuinely align with the intended use of the model.

\section{Conclusions}
\label{conclusions}

The evaluation of supervised machine learning models is far more than a mechanical step performed after training. As demonstrated throughout this paper, effective evaluation requires careful consideration of the relationship between the model, the data, the validation strategy, and the real objective of the task. A model that appears strong under a narrow or poorly chosen metric may, in practice, be unreliable, biased, or unsuitable for deployment. For this reason, evaluation should be treated as a central methodological component of machine learning rather than as a final reporting exercise.

Through the discussion of theoretical principles and a range of scenario-driven experiments, this study has shown that seemingly straightforward evaluation choices can have major consequences for how model performance is interpreted. In classification tasks, metrics such as accuracy may provide a misleading sense of success when datasets are imbalanced or when the costs of false positives and false negatives are not equal. In regression settings, aggregate scores such as RMSE, MAE, and $R^2$ each capture different aspects of model behavior and may lead to very different conclusions depending on the presence of outliers, skewed target distributions, or baseline effects. These findings reinforce the idea that no single metric can serve as a universal indicator of quality across all supervised learning problems.

The paper has also emphasized that evaluation errors often originate not only from metric misuse, but from weaknesses in the broader experimental design. Inappropriate train--test splits, overreliance on standard random cross-validation under dependent data conditions, and subtle forms of data leakage can all produce overoptimistic estimates of generalization performance. Such problems are especially serious because they may remain hidden until the model is exposed to real operational conditions. Accordingly, robust evaluation requires that the validation procedure be aligned with the structure of the data and the intended deployment setting, rather than selected solely for convenience or convention.

Another central conclusion of this work is that model evaluation is fundamentally decision-oriented. The practical value of a model does not depend only on statistical performance in the abstract, but on whether the evaluation criteria reflect the actual costs, risks, and objectives of the application domain. In medical diagnosis, for example, missing a positive case may be much more severe than generating a false alarm. In business prediction tasks, threshold selection and probabilistic calibration may be more important than maximizing overall accuracy. In scientific or policy-related contexts, trustworthy generalization and methodological transparency may matter as much as headline predictive performance. These considerations show that evaluation must be grounded in the real use case of the system.

Overall, the results of this study support a more careful and principled view of supervised machine learning evaluation. Practitioners should avoid relying on default metrics, resist the temptation to interpret single summary scores in isolation, and instead adopt a broader framework that combines multiple complementary metrics, appropriate validation strategies, and critical inspection of model errors. Doing so helps bridge the gap between apparent experimental success and genuine real-world reliability.

In conclusion, evaluation is not merely the final stage of the machine learning pipeline; it is the mechanism through which the validity, trustworthiness, and usefulness of a model are established. Sound evaluation practices are therefore essential not only for selecting better algorithms, but also for ensuring that machine learning systems are robust, transparent, and fit for purpose in the environments where they are ultimately used.

\section*{Acknowledgements}
For the purpose of open access, the authors have applied a Creative Commons Attribution (CC BY) licence to any Author Accepted Manuscript version arising from this submission.

\section*{Declarations}

\textbf{Funding}  
No funds, grants, or other support were received for this study.

\textbf{Competing interests}  
The authors have no relevant financial or non-financial interests to disclose.

\textbf{Authors' contributions}

\textbf{X. L.} contributed to the conceptualization and methodology of the study, curated the data, implemented the software, conducted the experimental evaluation, and wrote the original draft of the manuscript. \textbf{I. C. M.} contributed to the validation, and the review and editing of the manuscript. \textbf{M. T.} contributed to the conceptualization of the study, and participated in the review and editing of the manuscript. \textbf{X. X.} carried out the investigation, contributed to the formal analysis, and participated in the review of the manuscript. \textbf{N. P.} contributed to the conceptualization and methodology of the study, supervised the research, and participated in the review and editing of the manuscript.

\textbf{Data availability}  
The datasets analyzed in this study are publicly available from established online repositories. The corresponding dataset sources are cited in the manuscript, and the data can be accessed through the referenced repository links.

\textbf{Code availability}  
This study used publicly available datasets and the relevant references have been provided

\textbf{Ethics approval}  
This article does not contain any studies with human participants or animals performed by any of the authors.

\textbf{Consent to participate}  
Not applicable.

\textbf{Consent for publication}  
Not applicable.

\bibliography{References}
\end{document}